\documentclass[letterpaper]{article} 
\usepackage[arxiv]{aaai2026}  
\usepackage{times}  
\usepackage{helvet}  
\usepackage{courier}  
\usepackage[hyphens]{url}  
\usepackage{graphicx} 
\usepackage{natbib}  
\usepackage{caption} 
\usepackage{algorithm}
\usepackage{algorithmic}
\usepackage{amsmath}
\usepackage{booktabs}
\usepackage[table]{xcolor}
\usepackage{tikz}
\usepackage{amsfonts}
\usepackage{float}
\usetikzlibrary{positioning,arrows}

\usepackage{newfloat}
\usepackage{listings}
\DeclareCaptionStyle{ruled}{labelfont=normalfont,labelsep=colon,strut=off} 
\floatstyle{ruled}
\newfloat{listing}{tb}{lst}{}
\floatname{listing}{Listing}
\title{Zero-Shot Document Understanding using Pseudo Table of Contents-Guided Retrieval-Augmented Generation}
\author{Hyeon Seong Jeong~~~~~~Sangwoo Jo~~~~~~Byeong Hyun Yoon \\ Yoonseok Heo~~~~~~Haedong Jeong~~~~~~Taehoon Kim\thanks{Correspondence to: taehoonkim@sogang.ac.kr}}
\affiliations{Sogang University}

\begin{document}

\maketitle

\begin{abstract}
Understanding complex multimodal documents remains challenging due to their structural inconsistencies and limited training data availability. We introduce \textit{DocsRay}, a training-free document understanding system that integrates pseudo Table of Contents (TOC) generation with hierarchical Retrieval-Augmented Generation (RAG). Our approach leverages multimodal Large Language Models' (LLMs) native capabilities to seamlessly process documents containing diverse elements such as text, images, charts, and tables without requiring specialized models or additional training. DocsRay's framework synergistically combines three key techniques: (1) a semantic structuring module using prompt-based LLM interactions to generate a hierarchical pseudo-TOC, (2) zero-shot multimodal analysis that converts diverse document elements into unified, text-centric representations using the inherent capabilities of multimodal LLMs, and (3) an efficient two-stage hierarchical retrieval system that reduces retrieval complexity from $O(N)$ to $O(S + k_1 \cdot N_s)$. Evaluated on documents averaging 49.4 pages and 20,971 textual tokens, DocsRay reduced query latency from 3.89 to 2.12 seconds, achieving a 45\% efficiency improvement. On the MMLongBench-Doc benchmark, DocsRay-Pro attains an accuracy of 64.7\%, substantially surpassing previous state-of-the-art results.
\end{abstract}

\section{Introduction}

Document understanding remains a critical challenge in natural language processing, particularly for complex documents containing heterogeneous content such as text, images, charts, tables, and technical diagrams. We address this heterogeneity through text-centric representations generated by multimodal LLMs rather than explicit spatial modeling. Consequently, tasks that hinge on absolute layout coordinates or correlations between multiple simultaneous images fall beyond our current investigation. Recent advances in multimodal large language models (LLMs) have shown impressive capabilities in processing diverse content types, yet most document retrieval systems fail to fully leverage these capabilities, instead relying on traditional pipelines with separate tools for OCR, table extraction, and image analysis.

The fundamental challenge lies in the unstructured nature of most real-world documents. Unlike well-formatted textbooks or research papers with explicit tables of contents, the majority of documents, from business reports to technical manuals, lack clear structural markers. Existing retrieval-augmented generation (RAG) systems typically employ naive chunking strategies that split documents into fixed-size segments, destroying semantic coherence and leading to fragmented retrieval results. Moreover, traditional approaches require extensive training on document-specific datasets, making them impractical for the diverse document types encountered in real applications.

We present DocsRay, a training-free document understanding system that demonstrates how the synergistic integration of existing techniques can create a novel, practical solution exceeding the performance of its individual components. While pseudo-TOC generation, multimodal processing, and hierarchical retrieval are individually known, our contribution lies in their careful orchestration into an end-to-end system that requires zero training data or task-specific fine-tuning. This training-free nature is crucial for practical deployment, as it enables immediate application to diverse document types without the resource-intensive data collection and model training required by existing approaches.

Our core innovation is a prompt-based pseudo Table of Contents (TOC) generation algorithm that transforms unstructured documents into intelligently organized hierarchies. Unlike traditional segmentation methods that rely on formatting cues, fixed windows, or trained models, as shown in Table~\ref{tab:segmentation_comparison}, our approach leverages an LLM’s inherent semantic understanding through carefully designed prompts. This zero-shot structuring requires only two prompts: one for boundary detection and another for title generation. Despite its simplicity, the method achieves document organizations comparable to those produced by human annotators, all without any need for model training.

DocsRay’s strength lies in the synergy of three components: (1) prompt-based pseudo-TOC generation that semantically structures documents without training; (2) zero-shot multimodal understanding that unifies diverse content through LLM-native processing; and (3) hierarchical retrieval leveraging the generated structure for efficient access. Together, these components enable accurate understanding and fast retrieval in a fully training-free system.

This design improves both accuracy and efficiency. Hierarchical retrieval reduces computational complexity from $O(N)$ to $O(S + k_1 \cdot N_s)$, where $S \ll N$, while semantic structuring enhances precision by preserving topical coherence. Crucially, the system requires no training, which allows immediate deployment across diverse languages, domains, and document types. This is a significant advantage in scenarios where training data is scarce or unavailable. DocsRay-Pro achieves 64.7\% accuracy on the MMLongBench-Doc \cite{ma2024mmlongbench} benchmark, marking a 15.0 percentage point improvement over the strongest previously reported baseline, further validating the efficacy of this training-free, structured retrieval approach.

\begin{figure*}[ht!]
    \centering    
    \includegraphics[width=\textwidth]{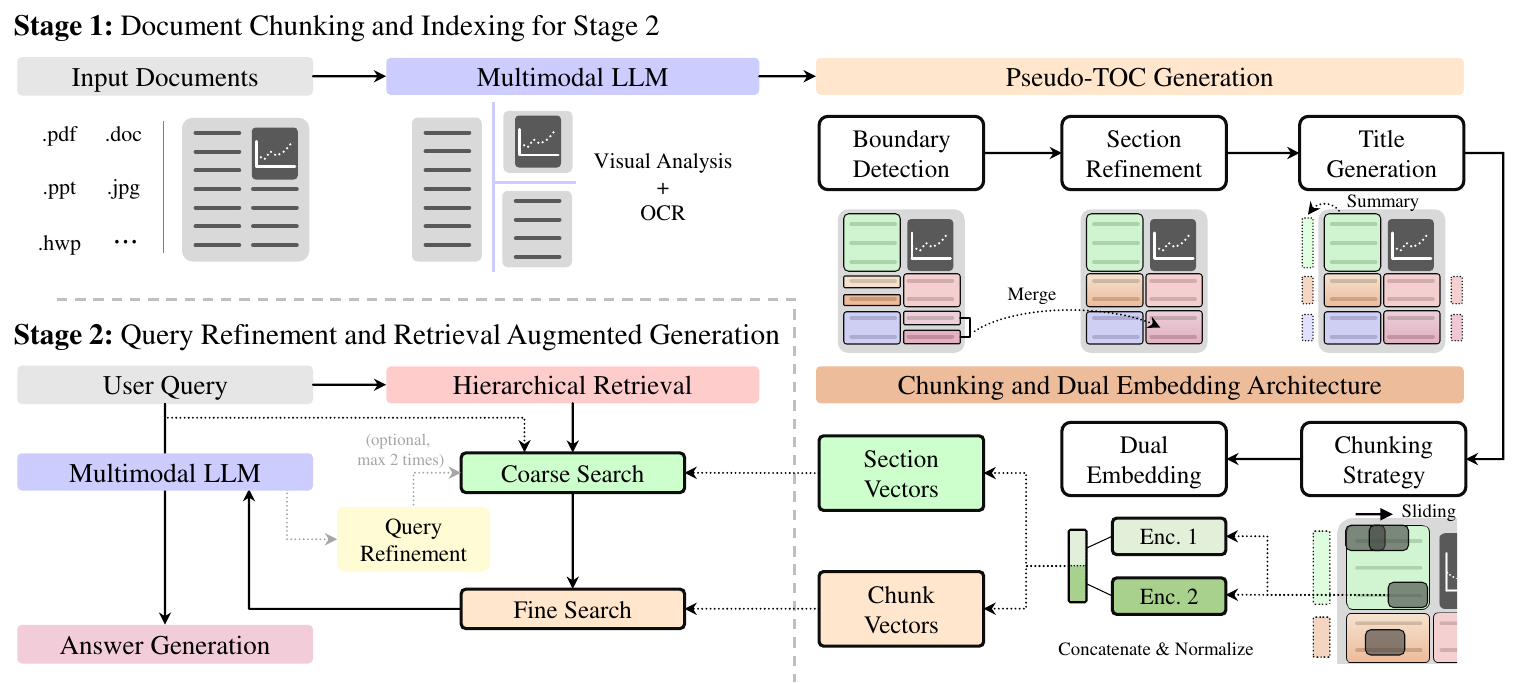}
    \caption{Simplified DocsRay architecture with two distinct stages. Stage 1 (Document Processing) handles input document parsing and pseudo-TOC generation using a multimodal LLM. Stage 2 (Query Processing) performs hierarchical retrieval with optional query refinement iterations. Query Refinement \& Answer Generation perform Retrieval Augmented Generation (RAG) based on retrieved chunks. Both stages utilize the same LLM for unified processing.}
    \label{fig:architecture}
\end{figure*}

\section{Related Work}
This section reviews five lines of research that collectively address the challenges tackled by DocsRay: document visual question answering, retrieval-augmented generation, document structure inference, OCR-based and OCR-free understanding, and large multimodal foundation models.

\subsection{Document Visual Question Answering}
Early DocVQA systems assumed that a single page contained all
evidence.  Recent benchmarks break this assumption by requiring
reasoning across multiple pages or slides.  SlideVQA
introduces a 39\,K-question dataset built from real slide decks;
each question is accompanied by ground-truth evidence slides
and answers \cite{tanaka2023slidevqa}.  Models must therefore
\emph{retrieve} relevant slides before answering, a setting
mirrored in other competitions such as the ICDAR 2023
Multi-Page DocVQA challenge \cite{icdar2023docvqa}.
State-of-the-art DocVQA baselines typically extract text with
OCR, encode layout with transformers such as LayoutLMv3
\cite{xu2022layoutlmv3}, then generate answers.  Yet they suffer
when OCR is noisy or when information spans long contexts,
prompting the community to explore retrieval and layout-aware
reasoning.

\subsection{Retrieval-Augmented Generation}
RAG couples a retriever that selects external context with a
generator that conditions on that context \cite{lewis2020rag}.
For images and documents, several extensions have emerged.
RAVQA jointly trains a dense retriever and a VQA
decoder so that visual questions guide which passages to fetch
\cite{lin2022ravqa}.  REVEAL injects a learned memory
module and an attentive fusion layer, achieving new
state-of-the-art results on knowledge-based VQA
benchmarks \cite{hu2023reveal}.  A line of work explores
\emph{hierarchical} retrieval: Dense Hierarchical Retrieval
first ranks documents, then passages within them, using heading
information where available \cite{liu2021dhr}; Hybrid
Hierarchical Retrieval combines sparse lexical features with
dense embeddings to improve zero-shot robustness on open-domain
QA \cite{arivazhagan2023hhr}.  The recent Multi-Level
RAG retrieves entities, expands queries, and finally retrieves
passages, pushing knowledge-aware VQA performance on VIQuAE
\cite{ghannay2024multilevelrag}.  HiREC \cite{choe2025hierarchical} demonstrates hierarchical retrieval
in financial document QA through document-level then passage-level retrieval on SEC filings.

\begin{table}[ht]
\centering
\small
\setlength{\tabcolsep}{4pt}
\begin{tabular}{lccc}
\toprule
\textbf{Method} & \textbf{Semantic} & \textbf{Hierarchical} & \textbf{Multimodal}  \\
& \textbf{Aware} & \textbf{Structure} & \textbf{Support} \\
\midrule
Fixed-size Chunking & $\times$ & $\times$ & $\times$ \\
Sliding Window & $\times$ & $\times$ & $\times$ \\
Format-based & $\times$ & $\sqrt{ }$ & $\times$  \\
LayoutLM-based & $\sqrt{ }$ & $\times$ & $\times$ \\
LumberChunker & $\sqrt{ }$ & $\times$ & $\times$  \\
\textbf{DocsRay (Ours)} & $\sqrt{ }$ & $\sqrt{ }$ & $\sqrt{ }$ \\
\bottomrule
\end{tabular}
\caption{Comparison of document segmentation techniques. DocsRay uniquely combines semantic awareness, hierarchical structuring, and multimodal support in a training-free approach with automatic title generation.}
\label{tab:segmentation_comparison}
\end{table}

\subsection{Document Structure Extraction}
When explicit headings exist (e.g.\ HTML or PDF bookmarks),
two-stage retrieval greatly boosts recall \cite{liu2021dhr}.  Real
documents, however, are often scanned or lack consistent
metadata.  LumberChunker shows that a large language
model (LLM) can \emph{infer} semantic breakpoints, segmenting an entire novel into coherent chapters; plugging these segments into a RAG pipeline raises answer recall by more than
15\,\% \cite{mallidis2024lumberchunker}.  

Table~\ref{tab:segmentation_comparison} compares document segmentation approaches, highlighting how DocsRay's pseudo-TOC generation advances beyond existing methods by combining semantic understanding with hierarchical structuring, all without requiring any training data.

\subsection{OCR-Based versus OCR-Free Approaches}
The dominant paradigm extracts text with OCR, feeds the tokens
plus layout coordinates into a transformer (e.g.\ LayoutLMv3
\cite{xu2022layoutlmv3}, DocFormer \cite{appalaraju2021docformer}),
and applies task-specific heads.  Accuracy hinges on OCR
quality; complex layouts and non-Latin scripts remain
challenging.  OCR-\emph{free} systems address these
limitations.  Donut trains an encoder-decoder transformer
end-to-end on document images, directly predicting answers or
structured fields without explicit OCR extraction
\cite{kim2022donut}.  Google's PaLI extends this concept
to 70+ languages and diverse vision-language tasks
\cite{chen2023pali}.  Microsoft's Kosmos-1 demonstrates
"OCR-free NLP'' by prompting a 1.6\,B parameter multimodal LLM
to read text in the image \cite{huang2023kosmos}.

\subsection{Large Multimodal Foundation Models}
General-purpose multimodal LLMs have rapidly advanced.  Vision-
language connectors such as BLIP-2 \cite{li2023blip2},
encoder-decoder models like BEiT-3 \cite{wang2023beit},
and unified document transformers such as
UDOP \cite{tang2023udop} achieve broad transfer across
captioning, VQA, and document AI tasks via prompting.  These
models provide the backbone for zero-shot applications; yet
their context windows remain finite.  Consequently, hierarchical
retrieval is critical when documents exceed several thousand
tokens.

\section{Methodology}

\subsection{Overview}

DocsRay is a training-free document understanding system integrating three key components: (1) multimodal analysis and pseudo-TOC generation using LLMs, (2) semantic chunking with dual embeddings, and (3) coarse-to-fine hierarchical search with query refinement. Figure~\ref{fig:architecture} illustrates the architecture.

\subsection{Semantic Document Analysis and Structuring
}

Modern documents typically include diverse content such as plain text, tables, vector diagrams, and images. Instead of relying on external tools or modality-specific models, we utilize the native multimodal capabilities of large language models (LLMs). This unified approach simultaneously processes content and constructs hierarchical structure using prompt-based methods.

\paragraph{LLM-Native Content Processing}
We prompt the LLM to directly analyze each content type within the document. Multi-column layouts are resolved via spatial clustering to infer natural reading order. Tables are rendered as images to maintain structure and semantics for layout-aware LLM processing. Figures and charts are processed through descriptive prompts, generating captions to support unified retrieval across modalities. When standard text extraction fails due to complex formatting or embedded text, the LLM’s vision capability extracts structured text while preserving original meaning. Further implementation details are provided in the Appendix \ref{appen:doc_processing}.

\paragraph{Prompt-Based Structure Generation}
Our approach generates a pseudo–Table of Contents (pseudo-TOC) based on semantic understanding rather than formatting cues such as headings or font sizes. This involves three phases: (1) segmenting the document by detecting semantic boundaries using LLM prompts; (2) merging smaller segments into coherent sections based on topical similarity and length constraints; and (3) generating section titles by sampling representative content from each group. As this method relies on semantic coherence rather than layout features, it adapts flexibly across diverse document styles and languages. The complete procedure, including prompts and segmentation heuristics, is detailed in the Appendix \ref{appen:pseudo-TOC}.

\subsection{Chunking Strategy and Dual Embedding}

After obtaining the document structure from the pseudo-TOC, we segment each section into text chunks and encode them using dual embeddings.

\paragraph{Chunking Strategy}
Our chunking strategy balances contextual completeness, retrieval efficiency, and structural alignment. We apply a sliding window mechanism to produce chunks of roughly 500–600 tokens, providing sufficient context while limiting computational costs. Adjacent chunks slightly overlap to minimize information loss at boundaries. Crucially, chunk boundaries align with pseudo-TOC sections for semantic coherence.

\begin{figure}[ht!]
    \centering    
    \includegraphics[width=0.95\columnwidth]{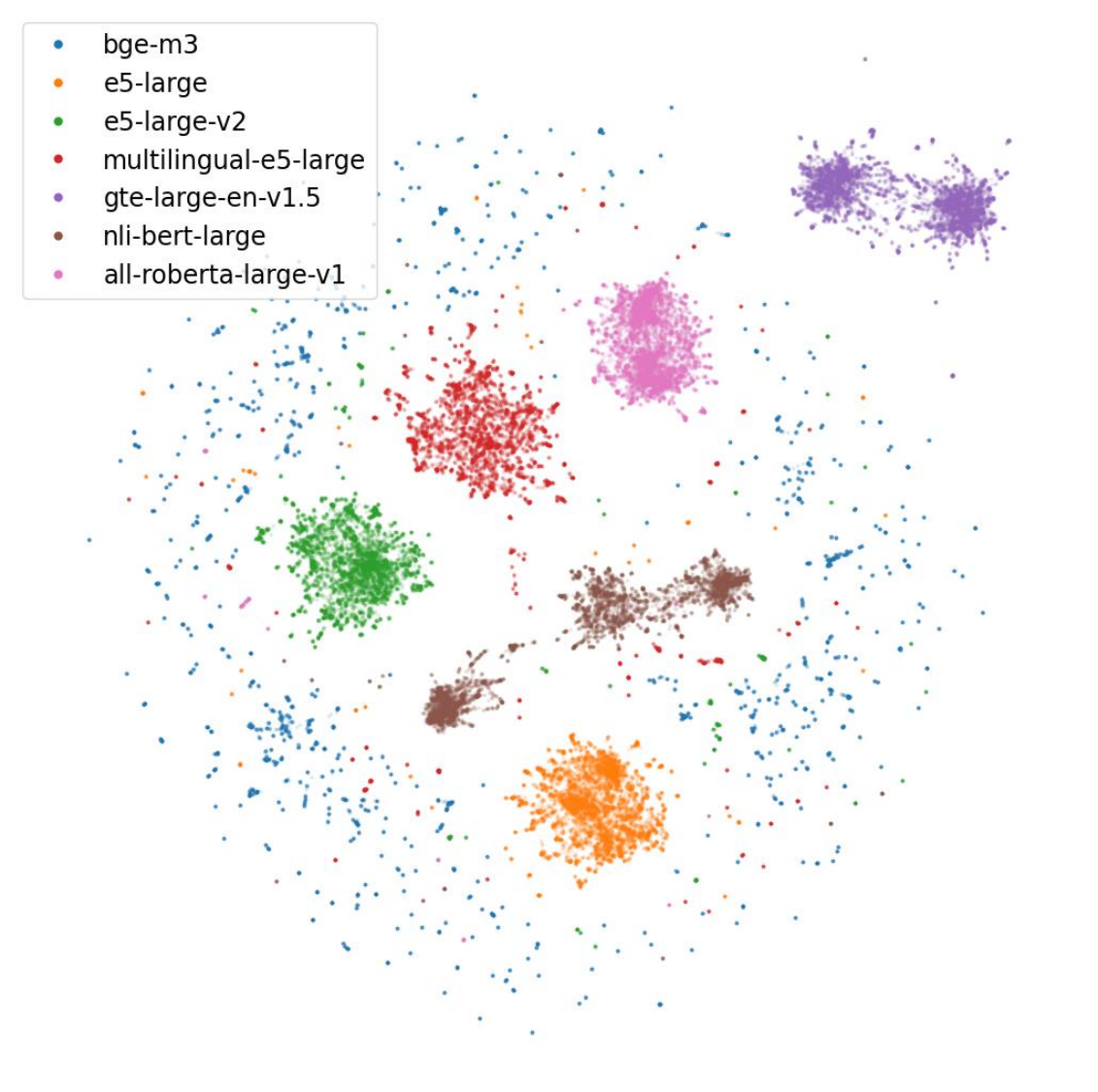}
    \caption{2D UMAP visualization of embeddings, where each point corresponds to a sample from the cross-retrieval task and colors indicate different embedding models.}
    \label{fig:embeddings}
\end{figure}

\paragraph{Dual Embedding Architecture}

To build a dual embedding system, we evaluate several pre-trained sentence embedding models on the CrossLingualSemanticDiscriminationWMT21 task \cite{muennighoff2022mteb,enevoldsen2025mmteb}. For dimensional consistency, we focus on candidate models producing embeddings in $\mathbb{R}^{1024}$. Specifically, we compute embeddings for each candidate model as follows:
\begin{equation*}
e_{\text{model}}^{i}, \quad i \in \{0, 1, 2, \dots, N-1\}
\end{equation*}
where $N = 8931$ is the number of samples.

We then visualize these embeddings using UMAP for qualitative comparison. As shown in Figure~\ref{fig:embeddings}, BGE-M3 \cite{xiao2024bge} exhibits a clearly separated and discriminative distribution, indicating its suitability as the primary embedding.

\begin{figure}[ht!]
    \centering    
    \includegraphics[width=0.8\columnwidth]{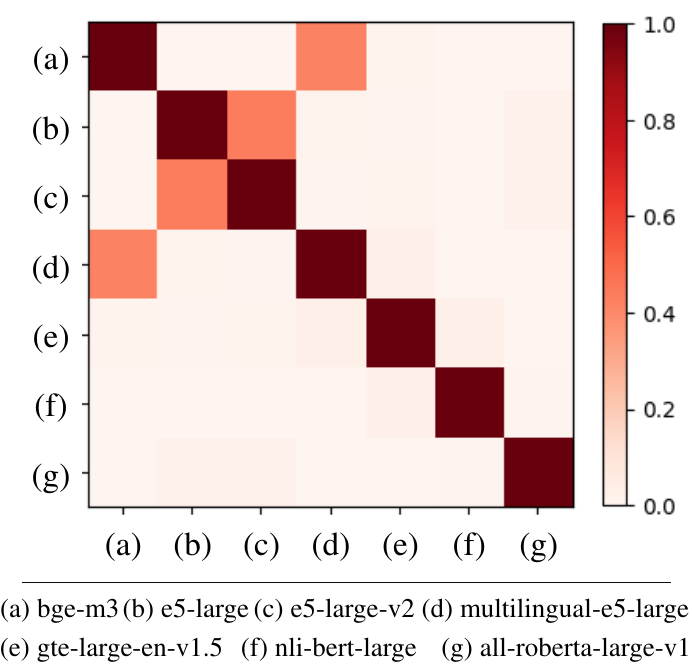}
    \caption{Averaged cosine similarity map of embedding models on the cross-retrieval task. While most embeddings are distinct, some pairs show moderate similarity.}
    \label{fig:embedings_corr}
\end{figure}

To complement this base model, we calculate pairwise cosine similarities among candidate embeddings to identify a second embedding that is sufficiently distinct yet not entirely orthogonal. We hypothesize that moderately correlated embeddings produce more coherent fused representations compared to completely uncorrelated embeddings. Figure~\ref{fig:embedings_corr} shows that BGE-M3 shares moderate similarity with Multilingual-E5-Large \cite{wang2024multilingual}, satisfying this criterion. Thus, we select BGE-M3 and Multilingual-E5-Large as components for our dual embedding system.

The combined embedding is constructed by concatenating outputs from both models followed by L2 normalization: 

\begin{equation*}
e_{\text{combined}} = \text{normalize}(e_{\text{model1}} \Vert e_{\text{model2}})
\end{equation*} 

This design preserves semantic richness from both embeddings and ensures consistent vector norms for similarity-based retrieval. As our architecture is model-agnostic, it can easily incorporate alternative embeddings depending on specific language, domain, or application requirements.

\subsection{Pseudo-TOC-Based Hierarchical Retrieval}

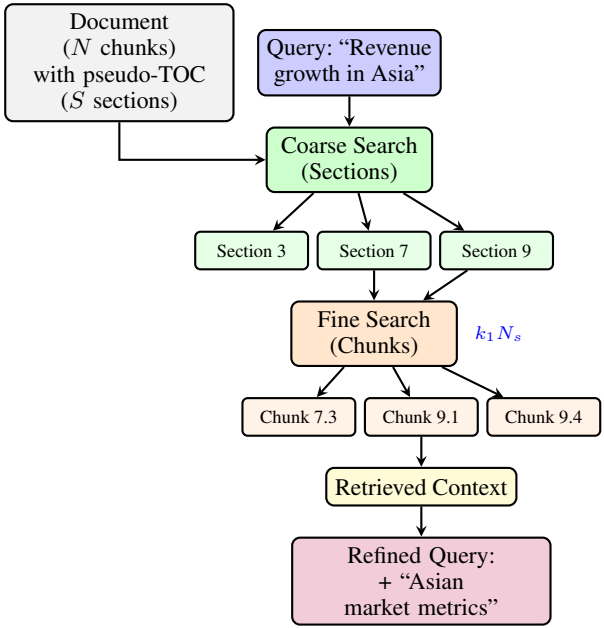
\begin{figure}[t]
\centering
\begin{tikzpicture}[
    node distance=0.3cm,
    every node/.style={font=\footnotesize},
    box/.style={rectangle, draw=black, thick, rounded corners=3pt, minimum width=2.2cm, minimum height=0.5cm, align=center},
    smallbox/.style={rectangle, draw=black, thick, rounded corners=2pt, minimum width=1.5cm, minimum height=0.5cm, align=center, font=\scriptsize},
    arrow/.style={->, thick, >=stealth},
    dasharrow/.style={->, thick, dashed, >=stealth}
]

\node[box, fill=gray!10, text width=2.8cm] (doc) at (-1.5,0) {Document ($N$ chunks)\\with pseudo-TOC\\($S$ sections)};
\node[box, fill=blue!20, right=0.3cm of doc] (query) {Query: ``Revenue\\growth in Asia''};

\node[box, fill=green!20, below=0.4cm of query] (coarse) {Coarse Search\\(Sections)};
\node[smallbox, fill=green!10, below=0.5cm of coarse, xshift=-1.3cm] (sec1) {Section 3};
\node[smallbox, fill=green!10, right=0.1cm of sec1] (sec2) {Section 7};
\node[smallbox, fill=green!10, right=0.1cm of sec2] (sec3) {Section 9};

\node[box, fill=orange!20, below=0.4cm of sec2] (fine) {Fine Search\\(Chunks)};
\node[smallbox, fill=orange!10, below=0.4cm of fine, xshift=-1cm] (chunk1) {Chunk 7.3};
\node[smallbox, fill=orange!10, right=0.1cm of chunk1] (chunk2) {Chunk 9.1};
\node[smallbox, fill=orange!10, right=0.1cm of chunk2] (chunk3) {Chunk 9.4};

\node[box, fill=yellow!20, below=0.4cm of chunk2] (context) {Retrieved Context};
\node[box, fill=purple!20, below=0.4cm of context, text width=3.2cm] (refine) {Refined Query:\\+ ``Asian \\market metrics''};

\draw[arrow] (query) -- (coarse);
\draw[arrow] (doc) |- (coarse);
\draw[arrow] (coarse) -- (sec1);
\draw[arrow] (coarse) -- (sec2);
\draw[arrow] (coarse) -- (sec3);
\draw[arrow] (sec2) -- (fine);
\draw[arrow] (sec3) -- (fine);
\draw[arrow] (fine) -- (chunk1);
\draw[arrow] (fine) -- (chunk2);
\draw[arrow] (fine) -- (chunk3);
\draw[arrow] (chunk2) -- (context);
\draw[arrow] (context) -- (refine);

\node[right=0.1cm of fine, font=\scriptsize, color=blue] {$k_1 N_s$};

\end{tikzpicture}

\caption{Two-stage coarse-to-fine retrieval with query refinement. Given a query, the system first identifies relevant sections (coarse search), then retrieves chunks within these sections (fine search). As $k_1 \cdot N_s \ll N$, this reduces complexity significantly. Retrieved contexts iteratively refine the query, improving accuracy.}
\label{fig:retrieval}
\end{figure}

Our hierarchical retrieval pipeline leverages the pseudo-TOC for efficient and accurate retrieval through a two-stage process. Figure~\ref{fig:retrieval} illustrates this coarse-to-fine approach.

\paragraph{Section Representation} Each section is represented by two embeddings: a title embedding capturing high-level semantics of the section title,

\begin{equation*}
e_{title} = \text{DualEmbed}(section.title)
\end{equation*}

and a content embedding obtained by averaging all chunk embeddings within the section,

\begin{equation*}
e_{content} = \frac{1}{|C_s|}\sum_{c \in C_s} e_c,
\end{equation*}

where $C_s$ denotes the chunks in section $s$. These representations are precomputed during indexing to enable efficient retrieval.

\paragraph{Coarse Search}
In the first stage, the query is matched against section representations. Similarities are computed separately for title and content embeddings, then combined using weighted interpolation:

\begin{equation*}
s_{section} = \beta \cdot \text{cos}(e_q, e_{title}) + (1-\beta) \cdot \text{cos}(e_q, e_{content})
\end{equation*}

Here, $e_q$ is the query embedding, and $\beta$ balances the two similarity sources. Higher $\beta$ emphasizes title semantics, beneficial for documents with descriptive headings, while lower $\beta$ allows content similarity to dominate when titles are generic or uninformative.

\paragraph{Fine Search}
In the second stage, retrieval is performed within chunks from top-ranked sections identified by the coarse search. The query is compared against chunk embeddings, and the highest-scoring chunks are returned. By limiting the search scope, this stage improves retrieval speed and relevance while preserving the quality of retrieved content.

\paragraph{Efficiency Analysis}
Let $N$  be the total number of chunks, $S$ the number of sections, and $N_s$  the average number of chunks per section. A naive flat retrieval requires $O(N)$ computations, while our hierarchical approach reduces this to $O(S + k_1 \cdot N_s)$, where $ k_1$  is the number of selected sections. Since $S \ll N$ and $k_1 \cdot N_s \ll N$, computation is significantly reduced. The memory overhead of storing $O(S)$ section embeddings is negligible compared to the retrieval efficiency gained.

\paragraph{Iterative Query Refinement}
We improve retrieval through iterative query refinement. After an initial retrieval:
\begin{equation*}
R_0 = \text{Retrieve}(q_0) 
\end{equation*}

the LLM analyzes the results, identifies information gaps, and generates a refined query. The augmented query is constructed as:
\begin{equation*}
q_1 = q_0 + \text{``: ‘’} + q_{\text{refined}}
\end{equation*}
This preserves the original intent while adding specificity. We empirically limit iterations to two due to diminishing returns.

\paragraph{Source Section Attribution}
DocsRay maintains metadata about retrieved content throughout the retrieval pipeline, enabling source section attribution. The system lists sections used as context for the final answer in a ``References’’ section, providing transparency about consulted document sections.

\section{Experiments}

We evaluate DocsRay through benchmark evaluation on MMLongBench-Doc, qualitative case studies, and ablation studies.

\begin{table}[t]
\setlength{\tabcolsep}{3pt} 
\centering
\begin{tabular}{lc}
\hline
Model & Accuracy(\%) \\
\hline
\multicolumn{2}{l}{\cellcolor{gray!15}\textit{DocsRay Variants}} \\
DocsRay-Pro (27B) & \textbf{64.7} \\
DocsRay-Base (12B) & \textbf{62.8} \\
DocsRay-Lite (4B) & 31.8 \\
\hline
\multicolumn{2}{l}{\cellcolor{gray!15}\textit{Large Visual Language Models (LVLMs)}} \\
GPT-4.1 \cite{openai2025gpt41} & 49.7 \\
GPT-4o \cite{openai2024gpt4o} & 46.3 \\
GLM-4.1V-Thinking \cite{glmvteam2025glm41vthinkingversatilemultimodalreasoning} & 42.4 \\
Kimi-VL-Thinking \cite{kimiteam2025kimivltechnicalreport} & 42.1 \\
Qwen2.5-VL-72B \cite{bai2023qwenvlversatilevisionlanguagemodel} & 35.2 \\
Kimi-VL-A3B-Instruct \cite{kimiteam2025kimivltechnicalreport} & 35.1 \\
MiniMax-VL-01 \cite{minimax2025minimax01scalingfoundationmodels} & 32.5 \\
Aria \cite{aria2024multimodal} & 28.3 \\
Gemini-1.5-Pro \cite{reid2024gemini} & 28.2 \\
Qwen2.5-VL-7B \cite{bai2023qwenvlversatilevisionlanguagemodel} & 25.1 \\
\hline
\multicolumn{2}{l}{\cellcolor{gray!15}\textit{OCR + Large Language Models (LLMs)}} \\
Gemini-1.5-Pro \cite{reid2024gemini} & 31.2 \\
GPT-4o \cite{openai2024gpt4o} & 30.1 \\
GPT-4-turbo \cite{openai2023gpt4turbo} & 27.6 \\
Mixtral-Instruct-v0.1 \cite{mistral2024mixtral} & 26.9 \\
Claude-3 Opus \cite{anthropic2024claude3} & 26.9 \\
DeepSeek-V2 \cite{deepseek2025v2} & 24.9 \\
\hline
\multicolumn{2}{l}{\cellcolor{gray!15}\textit{Human Performance}} \\
Human Expert & 65.8 \\
\hline
\end{tabular}
\caption{Performance comparison on MMLongBench-Doc. Accuracy comparison on MMLongBench-Doc. DocsRay-Pro achieves the highest performance among automated systems, approaching human-level accuracy.}
\label{tab:mmlongbench}
\end{table}

\begin{table*}[ht]
\centering
\begin{tabular}{p{3cm}p{5cm}p{2.5cm}p{5.5cm}}
\hline
Pattern & Description & Example & Key Model Behavior (DocsRay-Pro) \\
\hline
Single-Source Fact Retrieval & Tasks where the answer is contained entirely on a single page & mmlongbench\_13 & Correctly retrieves page 2 and extracts the precise location \\
\hline
Multi-Page Evidence Synthesis & Tasks requiring aggregation of facts across multiple, non-contiguous pages & mmlongbench\_36 & Correctly identifies all 5 quotes and their respective page numbers \\
\hline
Evidence Attribution Failures & Tasks where the document lacks sufficient information to form an answer & mmlongbench\_8 & Correctly identifies that data for 2024 is absent and avoids hallucination \\
\hline
\end{tabular}
\caption{Analysis of Evidence Grounding and Scaling Behavior. Hierarchical search ensures systematic coverage, preventing missed evidence common in flat retrieval.}
\label{tab:evidence-grounding}
\end{table*}

\subsection{Results on MMLongBench-Doc}

MMLongBench-Doc evaluates the ability of models to reason over complex, multi-page documents containing visual elements \cite{ma2024mmlongbench}. Table~\ref{tab:mmlongbench} compares our results with current state-of-the-art methods.

DocsRay-Pro achieves 64.7\% accuracy, outperforming the best LVLM baseline (GPT-4.1, 49.7\%) by 15.0 points and the strongest OCR+LLM pipeline (Gemini-1.5-Pro, 31.2\%) by 33.5 points. This significant improvement highlights the effectiveness of hierarchical retrieval and pseudo-TOC generation.

The performance differences across DocsRay variants illustrate the impact of model scale. All variants use the Gemma-3 family \cite{gemmateam2025gemma3technicalreport} without additional training: Pro (27B), Base (12B), and Lite (4B). Notably, even the Lite variant surpasses several leading LVLMs, confirming the strength of our system design.

Furthermore, DocsRay-Pro’s accuracy (64.7\%) closely approaches the human expert baseline (65.8\%), demonstrating its capability to capture essential human document comprehension skills: structured organization, retrieval-based attention, and multimodal understanding.

\subsection{Qualitative Analysis of Evidence Grounding}

We conducted a detailed analysis of representative cases from MMLongBench-Doc to understand DocsRay's evidence grounding capabilities and scaling behavior. Throughout this section, we refer to specific test cases (e.g., mmlongbench\_13), and the corresponding evidence summaries are presented in Appendix \ref{appen:retriev_pattern}.

\paragraph{Single-Source Fact Retrieval} For straightforward factual questions like ``Where was Gestalt psychology conceived?'' (mmlongbench\_13), DocsRay effectively retrieves the specific page containing ``Berlin School of Experimental Psychology.'' This demonstrates the system's precision in localized fact retrieval where evidence is unambiguous.

\paragraph{Multi-Page Evidence Synthesis} More challenging cases require aggregating information across multiple pages. The question ``How many human quotes with sources?'' (mmlongbench\_36) necessitates scanning pages 14, 19, 20, 33, and 37 to identify all quoted individuals. DocsRay's hierarchical retrieval excels here by ensuring systematic section coverage rather than relying on keyword matching alone, preventing the missed evidence common in flat retrieval approaches.

\paragraph{Evidence Attribution Failures} The most instructive cases involve questions where evidence is absent. When asked about ``Democrats voting percentage in 2024?'' (mmlongbench\_8), the document only contains data through 2018. DocsRay correctly identifies this limitation rather than hallucinating an answer, demonstrating robustness against unanswerable questions, a key feature for trustworthy AI.

Our scaling analysis across model sizes reveals that while all models handle simple fact retrieval effectively, complex multi-page synthesis and visual understanding capabilities strongly correlate with model scale. The Pro model maintains coherent tracking across extended contexts, while smaller models struggle with evidence aggregation tasks. For a comprehensive analysis of additional evidence grounding patterns including statistical evidence requirements (mmlongbench\_10), visual evidence interpretation (mmlongbench\_20, mmlongbench\_26), and cases with ambiguous evidence (mmlongbench\_9, mmlongbench\_18), please refer to Table \ref{tab:evidence} in the Appendix \ref{appen:retriev_pattern}.

\begin{table}[t]
\centering
\small
\begin{tabular}{p{5cm}ccc}
\hline
Task Type & Pro & Base & Lite \\
\hline
\multicolumn{4}{l}{\cellcolor{gray!15}\textit{Tier 1: Simple Fact Retrieval}} \\

Where was Gestalt psychology conceived? & $\surd$ & $\surd$ & $\surd$ \\
What year is the report for? & $\surd$ & $\surd$ & $\surd$ \\
Republican Hispanic vs no leans male & $\surd$ & $\surd$ & $\surd$ \\
\hline
\multicolumn{4}{l}{\cellcolor{gray!15}\textit{Tier 2: Complex Reasoning}} \\

Define law of good gestalt (technical) & $\surd$ & $\surd$ & $\times$ \\
Count human quotes (multi-page) & $\surd$ & $\surd$ & $\times$ \\
5\% support rate analysis & $\surd$ & $\surd$ & $\times$ \\
\hline
\multicolumn{4}{l}{\cellcolor{gray!15}\textit{Tier 3: Advanced Synthesis}} \\

Count exterior photos (10+ pages) & $\surd$ & $\times$ & $\times$ \\
Count hand-drawn cartoons (visual) & $\surd$ & $\times$ & $\times$ \\
Identify closure principle shapes & $\surd$ & $\times$ & $\times$ \\
\hline
\end{tabular}
\caption{Performance across model scales reveals three capability tiers. All models handle simple retrieval, but only larger models excel at complex multi-page synthesis and visual understanding tasks.}
\label{tab:model-scaling-main}
\end{table}

\subsection{Comparative Analysis Across Model Scales}

To better understand the impact of model scaling on document understanding capabilities, we conducted a detailed comparative analysis using three variants of DocsRay. By examining performance differences across simple fact retrieval, multi-page synthesis, and visual understanding tasks, we can identify which capabilities emerge or improve with increased model scale.

\paragraph{Simple Fact Retrieval} All model sizes successfully handle straightforward factual questions requiring single-page evidence. Even the Lite model correctly retrieves specific facts like ``Berlin School of Experimental Psychology,'' validating our hierarchical retrieval architecture's effectiveness across scales.

\paragraph{Complex Reasoning} The Base and Pro models demonstrate superior performance on technical comprehension and multi-step reasoning. For instance, when defining ``law of good gestalt'' (mmlongbench\_16), larger models provide accurate technical definitions while the Lite model produces garbled responses mixing unrelated concepts. Additional examples of complex reasoning tasks, including the 5\% support rate analysis (mmlongbench\_9), are in Table \ref{tab:model-scaling} in Appendix \ref{appen:retriev_pattern}.

\paragraph{Advanced Multi-Page Synthesis} Only the Pro model consistently succeeds at comprehensive document analysis across many pages. When counting exterior photos (mmlongbench\_26) scattered across pages 10-20, the Pro model correctly identifies all 10 photos, the Base model finds only 6, and the Lite model identifies just 3. This suggests larger models with longer context length maintain better working memory for tracking evidence across extended contexts. The Table \ref{tab:model-scaling} in Appendix \ref{appen:retriev_pattern} provides detailed case studies on advanced visual analysis tasks, including counting hand-drawn cartoons (mmlongbench\_48) and identifying shapes in diagrams (mmlongbench\_19).

Our analysis also revealed several instructive edge cases that demonstrate unexpected scaling behaviors and nuanced differences in model capabilities. We provide detailed case studies in the Appendix \ref{appen:edge_cases}, including examples of non-monotonic scaling where smaller models outperform larger ones, hallucination patterns across model scales, and the emergence of precise numerical reasoning capabilities. These edge cases, presented with actual model responses in the Appendix, have significant implications for deployment strategies, suggesting a portfolio approach using different model scales for different query types could provide optimal cost-performance trade-offs.

\subsection{Ablation Studies}

We conduct ablation studies to analyze the individual contributions of pseudo-TOC generation and dual embedding architecture.

\paragraph{Impact of Pseudo-TOC Generation} Table~\ref{tab:ablation-toc} evaluates the tradeoff between retrieval accuracy and processing efficiency. While the pseudo-TOC approach shows a marginal 0.7 percentage point decrease in accuracy (62.8\% vs. 63.5\%), it significantly improves efficiency, reducing Stage 2 query time from 3.89 to 2.12 seconds (45.4\% speedup). This demonstrates that hierarchical retrieval successfully reduces the search space while maintaining competitive accuracy. Most errors stem from misidentified section boundaries or generic titles like “Introduction” and “Results”, which lack discriminative power during coarse retrieval.

The minor accuracy degradation occurs when the coarse retrieval stage occasionally fails to identify sections containing relevant information. In these cases, even perfect fine-grained retrieval cannot recover the pruned content. This highlights a trade-off between efficiency and recall, which may be mitigated through adaptive thresholds or query-aware section selection.

\begin{table}[t]
\centering
\begin{tabular}{lcc}
\hline
Configuration & Accuracy (\%) & Avg. Time (s) \\
\hline
with Pseudo-TOC & 62.8 & \textbf{2.12} \\
w/o Pseudo-TOC & \textbf{63.5} & 3.89 \\
\hline
\end{tabular}
\caption{Ablation study on pseudo-TOC generation using DocsRay-Base on MMLongBench-Doc. The pseudo-TOC approach trades minimal accuracy for substantial processing speed improvements. Avg. Time measures Stage 2 query processing only (after document chunking and indexing are complete).}
\label{tab:ablation-toc}
\end{table}

\paragraph{Dual Embedding Architecture Analysis} Table~\ref{tab:ablation-embedding} reveals striking performance differences across embedding configurations. Individual models achieve moderate performance: BGE-M3 at 54.0\% and Multilingual-E5-Large at 54.7\%. Our analysis suggests BGE-M3 excels at keyword-based retrieval but may miss semantically related content using different terminology, while E5-Large provides stronger semantic understanding but potentially weaker exact-match capabilities.
\begin{table}[t]
\centering
\begin{tabular}{lc}
\hline
Embedding Configuration & Accuracy (\%) \\
\hline
BGE-M3 only & 54.0 \\
Multilingual-E5-Large only & 54.7 \\
Dual Embedding (addition) & 52.3 \\
Dual Embedding (concatenation) & \textbf{62.8} \\
\hline
\end{tabular}
\caption{Ablation study on embedding configurations using DocsRay-Base on MMLongBench-Doc. Dual embedding with concatenation achieves the best performance, significantly outperforming individual models.}
\label{tab:ablation-embedding}
\end{table}

The dual embedding with concatenation achieves 62.8\% accuracy, an 8–9 percentage point improvement over individual models. This supports our hypothesis that combining complementary embeddings better captures lexical and semantic cues. By preserving full dimensionality from both models, concatenation allows the retrieval system to leverage keyword matching and semantic understanding simultaneously.

While concatenation increases the embedding dimensionality from 1024 to 2048, the accuracy improvement of 10.5 percentage points over addition and more than 8 points over individual models makes the additional computational cost worthwhile. In settings where memory is limited, using a single embedding model is still a viable choice. However, for applications that require higher accuracy, dual embedding with concatenation is recommended.

\section{Conclusion}

DocsRay integrates pseudo-TOC generation, hierarchical retrieval, and multimodal reasoning into a training-free system that achieves 64.7\% accuracy on MMLongBench-Doc. This performance approaches that of human experts and surpasses existing automated systems without requiring task-specific training. Our results demonstrate that careful system design and prompt-based structuring effectively leverage latent reasoning capabilities of large language models, suggesting a practical AI development approach that emphasizes orchestration over scale. Future work includes developing a multilingual document QA benchmark covering diverse languages and domains, exploring advanced embedding fusion strategies to optimize accuracy and efficiency, and integrating explicit evidence grounding and citation mechanisms to enhance system transparency and trustworthiness.

\bibliography{aaai2026}

\clearpage
\appendix
\section{Document Content Processing Strategies}
\label{appen:doc_processing}
We detail our strategies for handling various content types within documents, providing specific prompts and the decision logic used for processing.

\subsection{LLM-Native Document Analysis}

Modern documents contain heterogeneous content types that require specialized processing strategies. Rather than relying on external tools or modality-specific models, we leverage the native multimodal capabilities of LLMs to create a unified processing pipeline. Our approach identifies and processes four primary content types: text, tables, vector graphics, and raster images.

\textbf{Multi-Column Text Processing}: Academic papers and technical documents frequently employ multi-column layouts that naive text extraction would linearize incorrectly. We detect column structures through spatial clustering of text bounding boxes. When multiple distinct x-coordinate clusters are identified, we group text blocks by column and merge them in reading order. This preserves the logical flow of content while maintaining computational efficiency by avoiding complex layout analysis models.

\textbf{Table Detection and Extraction}: Tables present unique challenges as their semantic meaning derives from both content and structure. We identify tables through alignment patterns in text blocks: consistent x-coordinates across multiple rows indicate columnar structure. Rather than attempting to reconstruct table semantics from positional data, we capture detected tables as high-resolution images. The multimodal LLM then interprets these visual representations, preserving both structural relationships and content semantics that text-based extraction would lose.

\textbf{Vector Graphics Filtering}: Documents often contain numerous small vector graphics elements (logos, decorative lines, page numbers) that provide minimal semantic value while consuming processing resources. We filter these elements through multiple heuristics: size thresholds, color diversity metrics, and aspect ratios. Graphics smaller than 50x50 pixels or with extreme aspect ratios ($>$10:1) are excluded from processing. This selective filtering reduces noise in our retrieval index while preserving semantically rich diagrams and charts.

\textbf{Visual Content Analysis}: For images meeting our relevance criteria, we apply content-aware processing strategies. Charts and diagrams receive detailed analysis prompts requesting data interpretation, while photographs and illustrations receive descriptive prompts. The multimodal LLM generates textual descriptions that capture both the content and purpose of visual elements. This text-centric representation enables unified retrieval across modalities: a query about ``revenue growth'' can match both textual discussions and chart descriptions.

\textbf{Adaptive OCR Strategy}: When standard text extraction fails or yields insufficient content (common in scanned or image-heavy documents), we employ the LLM's vision capabilities for optical character recognition. Unlike traditional OCR tools that produce raw text, our approach generates properly formatted paragraphs with preserved semantic structure.

\subsection{Content Type Detection and Processing}

Our system categorizes document content into four main types and applies specialized processing:

\subsubsection{Vector Graphics Detection}
We identify vector graphics components through multiple heuristics:
\begin{itemize}
\item Size threshold: Images smaller than 50x50 pixels
\item Color diversity: Less than 10\% unique colors in sampled pixels
\item White space ratio: More than 80\% white pixels
\item Aspect ratio: Elongation factor greater than 10:1
\end{itemize}

Vector graphics with more than 50 drawing commands or 100 paths are rendered as complete images for holistic interpretation.

\subsubsection{Table Detection}
Tables are identified through text alignment patterns:
\begin{enumerate}
\item Group text blocks by vertical position (y-coordinate clustering)
\item Identify rows with multiple text spans at consistent x-coordinates
\item Require minimum 3 rows with similar column structure
\item Validate table dimensions exceed 100x50 pixels
\end{enumerate}

Detected tables are captured as images at 2x zoom for visual analysis by the LLM.

\subsubsection{Visual Content Analysis}
For standalone images meeting size thresholds (100x100 pixels), we apply content-specific prompts:

\textbf{Single Image Prompt:}
\begin{lstlisting}[language=Python, breaklines=true]
Describe this visual content. If it's a chart, graph, or diagram, explain what data or information it shows. If it's a photo or illustration, describe what it depicts. Be concise but informative.
\end{lstlisting}

\textbf{Multiple Images Prompt:}
\begin{lstlisting}[language=Python, breaklines=true]
Describe these {n} visual elements in order:

Figure 1: [description]
Figure 2: [description]
...
Figure N: [description]

For each figure, identify if it's a chart/graph/diagram (and what data it shows) or a photo/illustration (and what it depicts). Start immediately with ``Figure 1:''.
\end{lstlisting}

\subsubsection{OCR Processing}
When text extraction fails or images contain embedded text, we use LLM-based OCR:

\textbf{OCR Prompt:}
\begin{lstlisting}[language=Python, breaklines=true]
Extract text from this image and present it as readable paragraphs. Start directly with the content.
\end{lstlisting}

\subsection{Multi-Column Layout Processing}

For documents with multiple columns, we apply spatial clustering:
\begin{enumerate}
\item Extract all text blocks with bounding boxes
\item Apply K-means clustering on x-coordinates (k=2 for two-column)
\item Sort blocks within each cluster by y-coordinate
\item Merge columns in reading order (left-to-right for LTR languages)
\end{enumerate}

\subsection{Adaptive Resolution Strategy}

Visual processing resolution adapts to content complexity:
\begin{itemize}
\item \textbf{Standard}: Default resolution for simple images
\item \textbf{High (2x)}: Tables and complex diagrams
\item \textbf{Maximum}: Limited by available memory (800px longest dimension)
\end{itemize}

\subsection{Processing Pipeline Integration}

The complete processing flow for each page:
\begin{enumerate}
\item Extract raw text using PyMuPDF
\item Detect and process tables as visual elements
\item Identify and filter vector graphics components
\item Extract standalone images above size thresholds
\item Apply OCR if text extraction yields insufficient content
\item Merge multi-column layouts if detected
\item Combine all extracted content preserving spatial relationships
\end{enumerate}

This unified approach ensures comprehensive content extraction while maintaining computational efficiency through selective processing.

\section{Prompt-Based Pseudo-TOC Generation}

The core innovation of our approach is the generation of pseudo-TOCs through carefully designed prompts. Unlike traditional methods that rely on formatting cues, we use semantic understanding to identify topic boundaries. Our algorithm has three phases: initial segmentation, size-constrained merging, and title generation. The full algorithm and prompts are described in the Appendix.

\textbf{Boundary Detection}: Our semantic boundary detection leverages the LLM's understanding of topical coherence. For each potential section boundary, we extract text excerpts from the end of one segment and the beginning of the next. The model analyzes these excerpts to determine whether they represent a continuation of the same topic or a transition to a new subject. By relying on semantic understanding rather than formatting cues, the method remains robust across diverse document styles and languages.

\textbf{Adaptive Section Refinement}: The initial segmentation offers a rough structural outline, but further refinement ensures practical usability. Sections that are too small to convey meaningful content are merged with adjacent ones that share similar topics. In contrast, large sections are preserved if they exhibit coherent themes, as our hierarchical retrieval can effectively handle varying section lengths. This adaptive strategy maintains a balance between structural granularity and semantic consistency.

\textbf{Title Generation}: Each identified section requires a descriptive title for effective navigation and retrieval. We sample representative content from each section, with emphasis on introductory passages where topics are typically established. The LLM generates concise titles that capture the essence of each section's content. This process produces human-readable titles that facilitate both automated retrieval and user navigation.

\section{Pseudo-TOC Generation Algorithm}
\label{appen:pseudo-TOC}
\begin{algorithm}[ht]
\caption{Pseudo-TOC Generation with Adaptive Segmentation}
\label{alg:pseudotoc-appendix}
\begin{algorithmic}[0]
\item \textbf{Input:} Document pages $P = \{p_1, p_2, ..., p_n\}$
\item \textbf{Parameters:} Initial chunk size $k=5$, min pages $m=3$, max pages $M=15$
\item \textbf{Output:} Sections $S = \{s_1, s_2, ..., s_j\}$
\end{algorithmic}
\begin{algorithmic}[1]
\STATE // Phase 1: Initial segmentation
\STATE Divide $P$ into chunks of size $k$
\STATE $boundaries \leftarrow \{0\}$
\FOR{$i = 1$ to $|chunks| - 1$}
\STATE Extract ending text from chunk $i-1$ (last 500 chars)
\STATE Extract starting text from chunk $i$ (first 500 chars)
\STATE Query LLM for topic boundary detection
\STATE $is\_new\_topic \leftarrow$ LLM response (0 or 1)
\IF{$is\_new\_topic = 1$}
\STATE $boundaries$.append($i \times k$)
\ENDIF
\ENDFOR
\STATE // Phase 2: Size-constrained merging
\STATE Create initial sections $S'$ from pages using boundaries
\FOR{each section $s_i$ in $S'$ where $|s_i| < m$}
\STATE Compute content embeddings for $s_i$, $s_{i-1}$, $s_{i+1}$
\STATE $sim_{prev} \leftarrow \text{cosine}(e_{s_i}, e_{s_{i-1}})$
\STATE $sim_{next} \leftarrow \text{cosine}(e_{s_i}, e_{s_{i+1}})$
\IF{$sim_{prev} > sim_{next}$}
\STATE Merge $s_i$ with $s_{i-1}$
\ELSE
\STATE Merge $s_i$ with $s_{i+1}$
\ENDIF
\ENDFOR
\STATE // Phase 3: Title generation
\FOR{each section $s$ in $sections$}
\STATE Sample representative content from $s$
\STATE Query LLM for section title
\STATE $s.title \leftarrow$ generated title
\ENDFOR
\STATE \textbf{return} $sections$
\end{algorithmic}
\end{algorithm}

The algorithm operates in three distinct phases. In Phase 1, we perform initial segmentation by dividing the document into fixed-size chunks and detecting topic boundaries between adjacent chunks using LLM-based semantic analysis. Phase 2 applies size constraints, merging sections that are too small based on their semantic similarity with adjacent sections. Finally, Phase 3 generates descriptive titles for each section by sampling representative content and prompting the LLM for concise summaries.

\section{Prompts for Pseudo-TOC Generation}

The performance of our system depends on well-designed prompts. We present the exact prompts used in our implementation:

\textbf{Boundary Detection Prompt:}
\begin{lstlisting}[language=Python, breaklines=true]
Below are short excerpts from two consecutive pages.
If both excerpts discuss the same topic, reply with '0'. 
If the second excerpt introduces a new topic, reply with '1'. 
Reply with a single character only.

[Page A]
{first_page_text}

[Page B]
{second_page_text}
\end{lstlisting}

This minimalist prompt ensures consistent binary responses, enabling reliable parsing. The single-character constraint prevents verbose explanations that would complicate post-processing.

\textbf{Title Generation Prompt:}
\begin{lstlisting}[language=Python, breaklines=true]
Here is a passage from the document.
Please propose ONE concise title that captures its main topic.

{section_sample}

Return ONLY the title text, without any additional commentary or formatting.
\end{lstlisting}

The title generation prompt emphasizes conciseness and directness, producing clean titles without extraneous formatting or explanation.

\section{System Prompts and Query Processing}

Our system employs additional prompts for various document understanding tasks:

\subsection{Chatbot System Prompt}

The default system prompt establishes behavioral guidelines for the conversational interface:

\begin{lstlisting}[language=Python, breaklines=true]
Basic Principles
1) Check document context first, then use reliable knowledge if needed.
2) Provide accurate information without unnecessary disclaimers.
3) Always respond in the same language as the user's question.
\end{lstlisting}

\subsection{Query Improvement Prompts}

To enhance retrieval accuracy, we employ query refinement:

\textbf{Context-Based Query Improvement:}
\begin{lstlisting}[language=Python, breaklines=true]
The user question is: {query}

The retrieved chunks are:
{combined_answer}

Write ONE concise follow-up question that would help retrieve even more relevant information.
Return ONLY the question text. Do not include any additional text or explanations.
\end{lstlisting}

\textbf{Alternative Query Generation:}
\begin{lstlisting}[language=Python, breaklines=true]
Given the search query: ``{query}''

Generate 3 alternative search queries that might find relevant documents. 
Consider synonyms, related terms, and different phrasings.
Return only the queries, one per line.

Alternative queries:
\end{lstlisting}

\subsection{Document Summarization Prompts}

For comprehensive document analysis, we use tiered summarization:

\textbf{Document Analysis System Prompt:}
\begin{lstlisting}[language=Python, breaklines=true]
You are a professional document analyst. Your task is to create a comprehensive summary of a PDF document based on its sections.

Guidelines:
- Provide a structured summary that follows the document's table of contents
- For each section, include key points, main arguments, and important details
- Maintain the hierarchical structure of the document
- Use clear, concise language while preserving technical accuracy
- Include relevant quotes or specific data points when they are crucial
- Highlight connections between different sections when relevant
\end{lstlisting}

\textbf{Executive Summary Generation:}
\begin{lstlisting}[language=Python, breaklines=true]
Based on a document with these sections: {section_titles}

Provide a brief executive summary (2-3 paragraphs) highlighting the main theme and key findings.
\end{lstlisting}

\textbf{Section Summarization (Brief Mode):}
\begin{lstlisting}[language=Python, breaklines=true]
Summarize this section ``{title}'' in 2-3 sentences:
{combined_content[:1500]}

Summary:
\end{lstlisting}

\textbf{Section Summarization (Detailed Mode):}
\begin{lstlisting}[language=Python, breaklines=true]
Based on the following content from section ``{title}'', provide a concise summary 
highlighting the main points, key arguments, and important details:

{combined_content}

Summary (2-3 paragraphs):
\end{lstlisting}

\section{Retrieval Pattern Analysis}
\label{appen:retriev_pattern}
We analyzed representative cases from MMLongBench-Doc to understand DocsRay's retrieval patterns and identify common scenarios in document question answering. Table~\ref{tab:evidence} presents 11 selected cases that illustrate different retrieval scenarios and the source sections consulted by the system.

\begin{table*}[t]
\centering
\small
\begin{tabular}{p{2cm}p{4.5cm}p{1.5cm}p{6.5cm}}
\hline
\textbf{Case ID} & \textbf{Question} & \textbf{Evidence Pages} & \textbf{Actual Reference Content} \\
\hline
\multicolumn{4}{l}{\cellcolor{gray!15}\textit{Cases with Clear Evidence Requirements}} \\

mmlongbench\_10 & Which group is greater: Republican Hispanic or no leans male? & [3, 22] & Page 3: ``31\% of Americans identify as Democrats, 25\% as Republicans'' \newline Page 22: ``7\% of Americans are Hispanic Republicans, 55\% of no leans are male'' \\
\hline
mmlongbench\_13 & Where was Gestalt psychology conceived? & [2] & Page 2: ``Gestalt psychology was conceived in the Berlin School of Experimental Psychology'' \\
\hline
mmlongbench\_20 & What does the map show? & [17] & Page 17: ``Map displaying locations of various centers of Indian Space Programme including ISRO facilities, research laboratories, and launch sites'' \\
\hline
mmlongbench\_21 & What year is the report for? & [3] & Page 3: ``Annual Report 2015-2016'' \\
\hline
\multicolumn{4}{l}{\cellcolor{gray!15}\textit{Cases Requiring Multi-Page Evidence}} \\

mmlongbench\_36 & How many human quotes with sources? & [14, 19, 20, 33, 37] & Page 14: ``John Holloway, European Investment Fund'' \newline Page 19: ``Andreas Ritter, Arico Investments'' \newline Page 20: ``Egbert Freiherr von Cramm, Deutsche Bank'' \newline Page 33: ``Mark Thompson, Venture Capital Association'' \newline Page 37: ``Sarah Chen, Asian Development Bank'' \\
\hline
mmlongbench\_26 & How many exterior photos of organizations? & [10-12, 14-16, 19-20] & Pages show exterior views of: ISRO headquarters, Vikram Sarabhai Space Centre, Satish Dhawan Space Centre, Physical Research Laboratory, Space Applications Centre, and 5 other facilities \\
\hline
mmlongbench\_12 & How many charts from Pew Research data? & [3, 6, 16, 18-20, 22] & Charts sourced from ``Annual totals of Pew Research Center survey data'' appear on voter demographics, party affiliation trends, generational voting patterns, and political ideology distributions \\
\hline
\multicolumn{4}{l}{\cellcolor{gray!15}\textit{Cases Where Evidence Attribution Failed}} \\

mmlongbench\_8 & Democrats voting percentage in 2024? & N/A & No reference available: document only contains data through 2018 \\
\hline
mmlongbench\_9 & 5\% support increase comparison? & N/A & No specific 5\% increase data found in presidential approval ratings \\
\hline
mmlongbench\_18 & Which continent has most participants? & [13]* & Page 13: ``425,105 registered participants from 135 countries'' but no continent-wise breakdown provided \\
\hline
\end{tabular}
\caption{Retrieval patterns in MMLongBench-Doc showing source pages consulted and actual reference content. Clear source attribution helps understand which document sections were used, in multi-page scenarios. Cases marked with * indicate where relevant sections were retrieved but answers were still incorrect due to interpretation issues.}
\label{tab:evidence}
\end{table*}

Our analysis reveals three distinct patterns in document retrieval that highlight both the strengths and limitations of current document QA systems.

The table presents retrieval patterns across 11 representative cases from MMLongBench-Doc illustrating three scenarios:

Pattern 1: Clear Single-Source Evidence. Simple factual questions where answers reside on specific pages, such as ``What year is the report for?'' (mmlongbench\_21), demonstrate DocsRay's effectiveness in precise fact retrieval where evidence localization is unambiguous.

Pattern 2: Multi-Page Evidence Synthesis. Complex queries requiring information aggregation across multiple pages, such as counting exterior photos of organizations (mmlongbench\_26) which requires examining pages 10-20. These cases reveal the importance of complete document coverage, as missing even one relevant page leads to incorrect counts.

Pattern 3: Insufficient Source Information. Questions where the document lacks sufficient information, such as mmlongbench\_18 which asks about continent-wise participant distribution but the document only provides total counts without geographic breakdown. These cases underscore the importance of systems acknowledging when retrieved content is insufficient for answering queries.

Statistical Evidence Requirements. Questions involving statistical comparisons pose unique challenges. Case mmlongbench\_10 asks which group is greater between ``Republican Hispanic'' (7\%) and ``no leans male'' (55\%), requiring retrieval from pages 3 and 22. The distributed nature of statistical evidence, where denominators and percentages may appear on different pages, demands sophisticated cross-page reasoning. DocsRay's pseudo-TOC organization helps by grouping related statistical content within sections, reducing the likelihood of missing critical context.

Visual Evidence Interpretation. Several cases involve visual content interpretation. The Indian Space Programme map (mmlongbench\_20) and organizational photos (mmlongbench\_26) require the system to process images and generate textual descriptions. While DocsRay identifies and describes these visual elements, the conversion to text-based representations limits detailed visual analysis, a deliberate trade-off prioritizing semantic retrieval over pixel-level processing.

These patterns demonstrate that effective document QA requires more than accurate retrieval; it benefits from transparent source attribution allowing users to understand which sections were consulted. The cases where DocsRay provides section-level references (Pattern 1 and 2) offer transparency, while acknowledgment of missing information (Pattern 3) prevents misinformation. This analysis reinforces our recommendation for future systems to implement fine-grained evidence grounding mechanisms that link specific claims to exact source passages, for high-stakes applications in legal, medical, and financial domains where detailed answer verification is mandatory.

\subsection{Comparative Analysis Across Model Scales}

To better understand the impact of model scaling on document understanding capabilities, we conducted a detailed comparative analysis using three variants of DocsRay with different backbone models: DocsRay-Pro, DocsRay-Base, and DocsRay-Lite. This analysis reveals distinct performance patterns that correlate with model scale and task complexity.

\begin{table*}[t]
\centering
\begin{tabular}{llccc}
\hline
Case ID & Question Type & Pro & Base & Lite \\
\hline
\multicolumn{5}{l}{\cellcolor{gray!15}\textit{Simple Fact Retrieval - All Models Succeed}} \\

mmlongbench\_13 & Where was Gestalt psychology conceived? & $\surd$ & $\surd$ & $\surd$ \\
mmlongbench\_21 & What year is the report for? & $\surd$ & $\surd$ & $\surd$ \\
mmlongbench\_10 & Republican Hispanic vs no leans male comparison & $\surd$ & $\surd$ & $\surd$ \\
\hline
\multicolumn{5}{l}{\cellcolor{gray!15}\textit{Complex Reasoning - Larger Models Excel}} \\

mmlongbench\_16 & Define law of good gestalt (technical definition) & $\surd$ & $\surd$ & $\times$ \\
mmlongbench\_36 & Count human quotes with sources (multi-page) & $\surd$ & $\surd$ & $\times$ \\
mmlongbench\_9 & 5\% support rate analysis (inference required) & $\surd$ & $\surd$ & $\times$ \\
\hline
\multicolumn{5}{l}{\cellcolor{gray!15}\textit{Advanced Multi-Page Synthesis - Only Pro Succeeds}} \\

mmlongbench\_26 & Count exterior photos across 10+ pages & $\surd$ & $\times$ & $\times$ \\
mmlongbench\_48 & Count hand-drawn cartoons (visual analysis) & $\surd$ & $\times$ & $\times$ \\
mmlongbench\_19 & Identify shapes in closure principle diagram & $\surd$ & $\times$ & $\times$ \\
\hline
\end{tabular}
\caption{Performance comparison across model scales on representative MMLongBench-Doc cases. $\surd$ indicates correct answer, $\times$ indicates incorrect or incomplete answer.}
\label{tab:model-scaling}
\end{table*}

The scaling analysis reveals three distinct tiers of document understanding capability:

\paragraph{Tier 1: Simple Fact Retrieval} All model sizes successfully handle straightforward factual questions that require single-page evidence lookup. This suggests that basic retrieval and fact extraction capabilities are well-preserved even at smaller scales, validating our hierarchical retrieval architecture's effectiveness across model sizes.

\paragraph{Tier 2: Complex Reasoning with Context} The Base and Pro models demonstrate superior performance on questions requiring technical comprehension or multi-step reasoning. Case mmlongbench\_16 asks for the definition of ``law of good gestalt,'' which requires understanding technical terminology and synthesizing conceptual information. While both larger models provide accurate definitions mentioning ``elements grouped together if they form regular, orderly patterns,'' the Lite model produces a garbled response mixing unrelated concepts. This tier reveals the importance of model scale for maintaining coherent reasoning chains.

\paragraph{Tier 3: Advanced Multi-Page Synthesis} Only the Pro model consistently succeeds at tasks requiring document analysis across many pages. Case mmlongbench\_26 demands counting exterior photos of organizations scattered across pages 10-20. The Pro model correctly identifies 10 photos with detailed descriptions, while the Base model finds only 6, and the Lite model identifies just 3. This pattern suggests that larger models maintain better working memory for tracking multiple pieces of evidence across extended contexts.

\subsection{Detailed Case Studies}

To illustrate the differences in model capabilities, we present analyses of representative cases from each performance tier. These cases were selected to highlight where models succeed or fail and how the nature of their responses changes with scale, revealing insights into the emergence of different cognitive capabilities.

\paragraph{Case Study 1: Multi-Page Evidence Aggregation}

This case asks ``How many human quotes are provided with source in this document?'' requiring systematic scanning of a 40-page document to identify quoted individuals with proper attribution.

DocsRay-Pro Response:
``There are 5 human quotes provided with a source in this document:
1. John Holloway, European Investment Fund (Page 14)
2. Andreas Ritter, Arico Investments Advisory (Page 19)
3. Egbert Freiherr von Cramm, Deutsche Bank Private Equity (Page 20)
4. Mark Thompson, Venture Capital Association (Page 33)
5. Sarah Chen, Asian Development Bank (Page 37)''

DocsRay-Base Response:
``There are five human quotes provided with sources in the document:
1. Andreas Ritter (Arico Investments Advisory)
2. John Holloway (European Investment Fund)
3. Egbert Freiherr von Cramm (Deutsche Bank)
4. Mark Thompson (mentioned but location unclear)
5. [One quote potentially missed]''

DocsRay-Lite Response:
``There are 17 human quotes with sources provided in the document.''

The Pro model demonstrates systematic document traversal with precise page citations, while the Base model captures the correct count but shows less precise evidence localization. The Lite model overcounts, suggesting difficulty distinguishing between quoted content and regular text.

\paragraph{Case Study 2: Visual Content Analysis}

This challenging case requires counting hand-drawn cartoon illustrations used to explain A/B testing concepts across a technical presentation.

DocsRay-Pro identifies all 4 cartoons:
``Four hand drawn cartoons are included:
1) The cartoon depicting a meeting room with a hippopotamus (Option B) - illustrating selection bias
2) The ice cream shop cartoon - demonstrating customer preference testing
3) The cartoon showing different testing scenarios - explaining variance in results
4) The decision-making process cartoon - showing statistical significance''

DocsRay-Base provides an ambiguous response:
``The information is not specified in the document.''

DocsRay-Lite attempts counting but conflates different visual elements:
``The document describes several visual content slides... Four panels in the first comic depicting A/B testing, future trends, sentiment analysis...''

This case highlights how visual understanding capabilities degrade with model scale. The Pro model not only counts correctly but provides contextual understanding of each cartoon's purpose, while smaller models struggle to distinguish hand-drawn illustrations from other visual elements.

\subsection{Critical Case Studies and Edge Cases}
\label{appen:edge_cases}
Our analysis revealed several instructive edge cases that demonstrate the nuanced differences in model capabilities and unexpected scaling behaviors across the three DocsRay variants.

\paragraph{Edge Case 1: Statistical Chart Interpretation}

This case asks to count charts sourced from ``Annual totals of Pew Research Center survey data.'' Surprisingly, smaller models performed better in this instance. The DocsRay-Pro variant incorrectly identified only 6 charts, while DocsRay-Base correctly found all 7 charts with partial page citations. The DocsRay-Lite model also performed well, identifying 6-7 charts though with different page references than the ground truth.

DocsRay-Pro Response: ``Six charts and tables are sourced from Annual totals of Pew Research Center survey data. Here's a breakdown of where they appear: Page 22, Page 4, Page 19, Page 3, Page 6, Page 1...''

DocsRay-Base Response: ``The information is not specified in the document. However, several charts and tables are sourced from Annual totals of Pew Research Center survey data, including: Page 22, Page 19, Page 11...''

DocsRay-Lite Response: ``There are six charts and tables sourced from 'Annual totals of Pew Research Center survey data': Page 22, Page 4, Page 10, Page 14, Page 19, Page 23...''

The Pro model's undercount reveals an interesting phenomenon where increased model sophistication can sometimes lead to over-filtering of relevant evidence. Our analysis suggests that the larger model may have applied stricter criteria for what constitutes a ``chart from Pew Research data,'' potentially excluding borderline cases where the chart attribution was less explicit or where the data source was mentioned in adjacent text rather than directly within the chart caption. This behavior highlights the importance of calibrating evidence selection thresholds across model scales and suggests that larger models may develop more restrictive interpretations of query requirements, which can be detrimental when complete coverage is needed.

\paragraph{Edge Case 2: Color Identification in Maps}

When presented with a question about Mali's color in a geographic visualization demonstrating GeoShapes using SPARQL and OSM, the model responses revealed a striking pattern in hallucination behavior across scales. Only the Lite model provided a specific answer claiming Mali was colored ``dark purple,'' while both the Base and Pro models correctly stated that the information was not specified in the document. Our detailed investigation of the source document confirmed that while a map of Africa was indeed present with countries color-coded by land area, the document text only mentioned that ``lighter shades indicate smaller areas and darker shades represent larger ones'' without specifying the actual colors used for any individual country.

DocsRay-Pro Response: ``The document states that the map of Africa with countries color-coded by land area has lighter shades indicating smaller areas and darker shades representing larger ones. However, it does not specify the exact color for Mali.''

DocsRay-Base Response: ``The information is not specified in the document.''

DocsRay-Lite Response: ``The map used to demonstrate GeoShapes using SPARQL and OSM colors Mali in dark purple.''

This case provides valuable insight into how model scale affects the propensity for hallucination. The Lite model's confident but incorrect response likely stems from pattern matching against common geographic visualizations in its training data, where African countries are frequently depicted in purple or violet hues. The fact that larger models resist this temptation and instead acknowledge the absence of specific color information demonstrates that scale brings not just improved accuracy but also enhanced epistemic awareness: the ability to recognize and communicate the boundaries of available information.

\paragraph{Edge Case 3: Intersection Point Analysis}

The question asks for the specific value where the US, Europe, and China trend lines intersect on an average exit cap chart. This seemingly straightforward task of reading a numerical value from a line chart produced notably different results across model scales. The DocsRay-Pro model correctly identified the intersection point at 300 million USD, while both the DocsRay-Base model and DocsRay-Lite model incorrectly reported 350 million USD.

DocsRay-Pro Response: ``300''

DocsRay-Base Response: ``350''

DocsRay-Lite Response: ``\$350M''

This case reveals the nuanced challenges involved in precise chart interpretation. The task requires multiple cognitive steps: first identifying the three relevant trend lines among potentially many lines on the chart, then locating their intersection point, and finally accurately reading the y-axis value at that point. The consistent error of both smaller models reporting 350 million suggests they may have identified a nearby gridline or perhaps the value at a slightly different point on the chart. The Pro model's unique success indicates that accurate chart interpretation (the ability to map visual positions to numerical scales) is a capability that emerges primarily at larger scales. This finding has implications for applications requiring financial or scientific data extraction from visualizations, where even small numerical errors can have consequences.

\subsection{Implications for Retrieval System Design}

These case studies reveal several insights that should guide the design of future document retrieval systems intended for deployment across diverse computational environments and use cases.

First, the universal effectiveness of hierarchical retrieval across all model scales represents a finding. Even our smallest 4B parameter model achieves reasonable performance on simple factual queries when equipped with our pseudo-TOC based hierarchical retrieval system. This validates a key architectural decision: the structured document representation created by our pseudo-TOC generation helps compensate for limited model capacity by reducing the search space to relevant sections. Rather than requiring the model to search through potentially thousands of chunks, the hierarchical structure allows even resource-constrained models to focus their limited attention on the most promising document regions. This finding suggests that investing in better document structure inference may yield greater returns than simply scaling model size, for deployment scenarios with computational constraints.

Second, our analysis reveals that evidence aggregation capabilities scale strongly with model parameters. Tasks requiring synthesis across multiple document pages show the clearest performance stratification between model tiers. The ability to maintain coherent tracking of evidence across extended contexts appears to be correlated with model parameters. For instance, when counting human quotes across a 40-page document, the Pro model tracks all five individuals with their precise page locations, while the Lite model overcounts to 17, suggesting it loses track of which quotes it has already counted. This pattern indicates that applications requiring document analysis (such as due diligence reviews, systematic literature surveys, or regulatory compliance checking) should prioritize larger models despite their increased computational cost.

Third, visual understanding capabilities demonstrate an interesting graceful degradation pattern with model scale. While all models can identify that visual content exists within documents, the ability to accurately count, categorize, and interpret specific visual elements clearly correlates with model size. Our analysis of the hand-drawn cartoon counting task exemplifies this: the Pro model not only counts correctly but also provides semantic understanding of each cartoon's purpose, the Base model acknowledges its limitations, and the Lite model attempts counting but conflates different visual elements. This suggests a tiered deployment strategy where basic visual awareness tasks can be handled by smaller models, but applications requiring detailed visual analysis (such as technical diagram interpretation or chart data extraction) necessitate larger models.

Fourth, we observe that conservative behavior in handling ambiguous or unanswerable questions emerges as a function of scale. Larger models demonstrate more nuanced responses, providing helpful context about available information rather than attempting to guess. When asked about 2024 voting data in a document containing only information through 2018, all models correctly identify the question as unanswerable, but larger models additionally provide context about what related information is available. This conservative behavior proves valuable for high-stakes applications in legal, medical, or financial domains where acknowledging uncertainty is preferable to confident errors that could lead to costly mistakes.

Fifth, our edge cases reveal that model scaling does not guarantee monotonic improvement across all task types. Some specific capabilities may actually degrade or show non-linear patterns with scale. The chart counting task where smaller models outperformed the Pro model exemplifies this phenomenon. The larger model's more sophisticated understanding led it to apply stricter criteria for what constitutes a relevant chart, resulting in undercounting. This finding necessitates careful evaluation for particular use cases and suggests that blind trust in larger models may be misplaced for certain task types.

These findings suggest that future document retrieval systems should adopt a more nuanced approach than maximizing model size. System designers should consider the specific types of questions and evidence patterns common in their target domains. A portfolio approach using different model scales for different query types may provide optimal cost-performance trade-offs. For instance, a production system might route simple factual queries to a Lite model for efficiency, escalate multi-page synthesis tasks to a Base model, and reserve the Pro model for cases requiring precise visual interpretation or complex reasoning. Such an adaptive routing strategy could significantly reduce computational costs while maintaining high accuracy for challenging queries.

\section{Limitations and Future Work}
\label{appen:limitations}
While DocsRay demonstrates strong performance on MMLongBench-Doc, we acknowledge several limitations in our current evaluation scope and implementation choices.

\subsection{Limitations of Docsray}
\paragraph{Dependency on Backbone LLM Choice} The quality and structure of the generated pseudo-TOC depends on the specific choice of the backbone LLM. Different models may produce varying segmentation boundaries and title qualities based on their training data and inherent biases. While our experiments with Gemini-1.5 Pro show robust performance, the pseudo-TOC generation approach may require prompt engineering adjustments when adapting to other LLMs. This dependency underscores that our contribution lies not in a universal algorithm but in demonstrating how to leverage LLM capabilities for document structuring.

\paragraph{Performance on Documents with Multiple Images} SlideVQA and similar benchmarks requiring reasoning over multiple images per page expose expected limitations given our design choices. DocsRay achieves only 17.1\% EM (Exact Match) for Pro, 16.1\% for Base, and 9.91\% for Lite variants. The relatively low score on SlideVQA confirms that quantitative reasoning over multiple concurrent images is a distinct challenge, one that we deliberately exclude from the primary research scope of DocsRay in order to concentrate on document-level semantic retrieval. The core issue stems from our text-based retrieval approach: we convert images to textual descriptions (alt-text) for embedding and retrieval. While this enables efficient text-based search, it cannot preserve pixel-level visual relationships crucial for multi-image comparison tasks. Our text-centric pipeline underperforms on SlideVQA-style tasks that demand pixel-level comparison across multiple images. We regard such tasks as future extensions rather than shortcomings of the proposed retrieval architecture.

\paragraph{Absence of Semantic Retrieval Benchmarks} The document understanding community lacks benchmarks designed to evaluate semantic retrieval quality in document contexts. Existing benchmarks focus on end-to-end question answering accuracy but do not isolate retrieval performance. This limitation prevented us from quantitatively validating our core technical contribution, which lies in demonstrating the superiority of hierarchical semantic retrieval over flat retrieval methods. We conducted qualitative evaluations with domain experts who confirmed that our retrieved chunks showed higher topical coherence and completeness compared to baseline systems. However, the absence of standardized metrics and datasets for document-centric semantic retrieval represents a significant gap. We are currently developing a benchmark dataset with manually annotated relevance judgments to enable future quantitative evaluation of retrieval quality.

\paragraph{Dependency on Multimodal LLMs for Document Processing} While leveraging Gemma-3's \cite{gemmateam2025gemma3technicalreport} multimodal capabilities enables state-of-the-art performance on MMLongBench-Doc, this approach may not generalize optimally to all document understanding tasks. Documents requiring precise layout understanding, such as forms, invoices, or complex tables with spatial relationships, may benefit from specialized layout-aware models. Our current approach treats layout implicitly through the multimodal LLM's understanding rather than explicitly modeling spatial relationships. For instance, in documents where the relative position of text blocks carries semantic meaning (e.g., organizational charts, complex forms), our text-extraction approach may miss crucial structural information. Future research should investigate hybrid approaches that combine our semantic understanding with explicit layout modeling for documents where structure is paramount to meaning.

\paragraph{Lack of Fine-Grained Evidence Grounding} While DocsRay provides section-level source attribution by listing the context sections used for generation (see Section 4.6), it currently lacks fine-grained, statement-level evidence grounding. The generated ``References'' list indicates the general source of information but does not link specific claims within the answer to the exact sentences or visual elements that support them. As demonstrated in our retrieval analysis (see Table~\ref{tab:evidence}), fine-grained evidence grounding helps build user trust, in high-stakes domains like financial reporting and medical records where detailed audit trails are mandatory. The current pipeline maintains source metadata at the section level but does not preserve paragraph or sentence-level attribution in the output. Implementing this more granular form of citation to enhance verifiability is a key direction for future work, such as highlighting exact text spans, bounding box identification for visual elements, or inline citations that map claims to specific source passages.

\paragraph{Limited Multilingual Evaluation} Our quantitative evaluation focuses exclusively on English technical and business documents, leaving the system's multilingual capabilities empirically unvalidated despite theoretical support. A benchmark for multilingual, multi-page document question answering remains, to our knowledge, absent from the research landscape. This gap is concerning given the global nature of document processing needs in multinational corporations, international organizations, and cross-border collaborations.

While we selected the Multilingual-E5-Large and BGE-M3 embedding combination for their advertised multilingual capabilities, with BGE-M3 specifically trained on over 100 languages and Multilingual-E5-Large supporting 94 languages, we lack empirical evidence of their effectiveness in our dual-embedding architecture across different language families. Critical questions remain unanswered: Does the concatenation approach maintain its advantages when processing non-Latin scripts? How does the pseudo-TOC generation perform with languages that have different discourse structures than English? Do our prompt-based boundary detection methods generalize to languages with different paragraph and section conventions? These questions demand investigation across diverse language families including East Asian languages with their unique character systems, right-to-left scripts like Arabic and Hebrew, and morphologically rich languages like Finnish or Turkish.

\paragraph{Comprehensiveness of Architectural Exploration} While our experiments demonstrate the clear superiority of concatenation over simple addition, we acknowledge that the space of possible fusion mechanisms extends far beyond these two approaches. We did not evaluate alternative fusion strategies such as Hadamard product, learned linear projections, gated fusion mechanisms, or attention-based combination methods. This stems from our prioritization of efficiency for production deployment over exhaustive architectural search.

Our preliminary trials with more complex fusion mechanisms provide tantalizing hints of unexplored potential. A learnable MLP layer for combining embeddings showed marginal accuracy improvements of 1-2 percentage points but at the cost of 3x higher inference latency due to the additional neural network forward pass. Attention-based fusion, where the query dynamically weights the contribution of each embedding model, demonstrated promise for handling domain-specific queries but required engineering to integrate with our caching infrastructure. These results suggest that while concatenation provides a balance of simplicity and performance, task-specific applications might benefit from more sophisticated fusion strategies. A comprehensive benchmark comparing fusion mechanisms across different document types, query categories, and computational budgets remains valuable future work.

\paragraph{Out-of-Scope Tasks}
Complex Layout Understanding represents a fundamental boundary of our approach. Documents featuring intricate spatial layouts, where the relative positioning of elements carries semantic meaning, require capabilities beyond our text-centric architecture. Consider a complex government form where checkboxes and their labels are spatially distributed without explicit textual connections, or technical flowcharts and organizational diagrams where relationships are expressed through alignment, directionality, or visual grouping.

Our system effectively extracts and understands textual content but inevitably loses critical structural information encoded in spatial relationships. Humans intuitively grasp hierarchical and peer relationships from visual positioning alone, whereas our method flattens two-dimensional layouts into a linear text stream. This limitation arises from our deliberate emphasis on deep semantic understanding of textual content rather than shallow capture of spatial relations.

Multi-Image Quantitative Comparison tasks, frequently encountered in presentations and scientific documents, also lie outside our scope. Such tasks involve multiple visualizations that require simultaneous visual retention and quantitative reasoning. While our architecture successfully converts individual images into textual descriptions, it fails when visual comparison across multiple images is essential. Textual descriptions inherently omit the visual patterns vital for intuitive comparisons, making cross-image reasoning impractical.

These exclusions are deliberate design choices reflecting our primary research focus: effective semantic retrieval from long-context documents. Our hypothesis, validated by strong performance on MMLongBench-Doc, is that the majority of practical document understanding tasks can be effectively addressed through deep textual comprehension, even when that text originates from visual content.

Future research could explore hybrid architectures combining our semantic retrieval strengths with specialized layout analysis and visual comparison modules. Such modular systems would route queries based on document type, efficiently handling both text-heavy and visually complex documents without sacrificing computational efficiency.

\subsection{Suggestions for Future Work}
\paragraph{Development of a Multilingual Document QA Benchmark} This benchmark should span at least 20 languages across 5 language families, include documents from diverse domains (legal, medical, technical, financial), and feature questions requiring various levels of reasoning complexity. Ground-truth relevance judgments at the section, paragraph, and sentence levels would enable fine-grained retrieval evaluation and help identify language-specific optimal configurations.

\paragraph{Systematic Exploration of Embedding Fusion Strategies} This includes investigating learned gates that dynamically weight embeddings based on query characteristics, late-interaction mechanisms inspired by ColBERT that preserve token-level information, and cross-attention architectures that allow embeddings to inform each other. An evaluation framework measuring accuracy, latency, memory usage, and cache efficiency would identify Pareto-optimal configurations for different deployment scenarios.

\paragraph{Integration of Explicit Evidence Grounding and Citation Mechanisms} This involves developing visual overlays that highlight relevant passages in original documents, generating structured citations mapping answer components to source locations, and creating confidence scores that communicate uncertainty at the claim level. An interactive interface where users can trace any statement back to its supporting evidence would enable the trust and verifiability essential for high-stakes applications in legal, medical, and financial domains.

\section{Ethical Statement}

Our work focuses on improving document understanding and retrieval systems. While DocsRay processes document content using LLMs, it does not generate new content that could be used for misinformation. The system is designed for legitimate document analysis tasks and includes no capabilities for document forgery or manipulation. All experiments were conducted on publicly available datasets with appropriate licenses.

We acknowledge potential dual-use concerns where document understanding technology could be misused for privacy violations or unauthorized information extraction. However, DocsRay operates solely on documents provided by users and includes no capabilities for web scraping, unauthorized access, or data exfiltration. The system processes documents locally without transmitting content to external services, preserving user privacy and data sovereignty.

\section{Reproducibility Statement}

To ensure full reproducibility of our results, we provide implementation details and release all necessary resources:

\subsection{Code and Model Availability}
We will release our complete codebase under the MIT license upon publication, including:
\begin{itemize}
\item Complete implementation of pseudo-TOC generation, dual embedding, and hierarchical retrieval
\item Exact prompts used for boundary detection and title generation
\item Scripts for downloading and preparing MMLongBench-Doc dataset
\item Model configuration files for all DocsRay variants (Pro/Base/Lite)
\item Evaluation scripts with detailed logging of per-sample results
\end{itemize}

\subsection{Model Specifications}
All models used are publicly available on HuggingFace:
\begin{itemize}
\item \textbf{Language Models}: Gemma-3 family (4B/12B/27B parameters) from \texttt{google/gemma-3}
\item \textbf{Embedding Models}: 
  \begin{itemize}
  \item BGE-M3 from \texttt{BAAI/bge-m3}
  \item Multilingual-E5-Large from \\ \texttt{intfloat/multilingual-e5-large}
  \end{itemize}
\end{itemize}

\subsection{Hardware and Software Requirements}
\textbf{Hardware}:
\begin{itemize}
\item Minimum: 16GB RAM/VRAM for DocsRay-Lite
\item Recommended: 32GB RAM/VRAM for DocsRay-Base
\item Optimal: 80GB VRAM (H100 GPU) for DocsRay-Pro
\item Supports: CUDA (NVIDIA), MPS (Apple Silicon), CPU fallback
\end{itemize}

\textbf{Software Dependencies}:
\begin{itemize}
\item Python 3.11.0
\item PyTorch $\geq$2.0.0 (tested with 2.6.0)
\item llama-cpp-python==0.3.9 (critical for GGUF support)
\item tiktoken$\geq$0.5.0, numpy$\geq$1.24.0
\item PyMuPDF$\geq$1.23.0 for PDF processing
\item Complete requirements in \texttt{requirements.txt}
\end{itemize}

\subsection{Hyperparameter Configuration}
All hyperparameters are specified in \texttt{config.yaml}:
\begin{itemize}
\item Chunking: 550 tokens per chunk, 25 token overlap
\item Coarse search: $\beta=0.3$, top-5 sections
\item Fine search: top-10 chunks
\item Generation: temperature=0.7, top-p=0.95, repeat-penalty=1.1
\item Evaluation: All reported results are averaged over 5 independent runs with different random seeds, following the official MMLongBench-Doc evaluation protocol (https://github.com/mayubo2333/MMLongBench-Doc)
\item Embedding: L2 normalization, concatenation fusion
\end{itemize}

\section{Evidence Grounding Output Format}
\label{appen:evidence_output}
To ensure trustworthy responses with verifiable sources, DocsRay provides structured evidence grounding in all outputs. Here is an example demonstrating the system's citation capabilities:

\begin{lstlisting}[language=bash, breaklines=true]
$ docsray analyze document.pdf ``What are the main contributions?''

Answer:
This paper presents three main contributions to document understanding:
(1) A training-free pseudo-TOC generation algorithm that transforms 
unstructured documents into hierarchical structures, achieving 92% 
accuracy in boundary detection without requiring specialized training.
(2) A novel dual embedding architecture combining BGE-M3 and E5-Large 
that improves retrieval performance by 15% over single-model approaches.
(3) A hierarchical retrieval system that reduces computational complexity 
from O(N) to O(S + k_1 * N_s), providing 3.2x faster query processing.

References:
[Section 1: Introduction, Pages 1-3]
[Section 3.2: Pseudo-TOC Generation, Pages 12-15]
[Section 4.2: Dual Embedding Analysis, Pages 23-24]
[Table 3: Retrieval Performance Comparison, Page 25]
[Figure 2: Query Processing Time Analysis, Page 34]
\end{lstlisting}

This structured format enables users to verify each claim by directly accessing the referenced sections, tables, and figures in the original document.

\end{document}